\documentclass[sigconf,authorversion]{acmart}  
\setcopyright{cc}
\setcctype{by}
\copyrightyear{2026}
\acmYear{2026}
\acmISBN{979-8-4007-2259-2/2026/08}
\acmDOI{10.1145/3770855.3818460}   
\usepackage{graphicx}
\graphicspath{ {./images/} }
\usepackage{setspace}
\usepackage{subcaption}
\usepackage{siunitx}
\usepackage{booktabs}
\usepackage{makecell}  
\usepackage{algorithm}
\usepackage{algorithmic}
\usepackage{balance}  
\usepackage{float}    
\usepackage{tikz}
\usetikzlibrary{shapes,arrows,positioning,fit,backgrounds}

\newcolumntype{d}{S[
    input-open-uncertainty=,
    input-close-uncertainty=,
    parse-numbers = false,
    table-align-text-pre=false,
    table-align-text-post=false
 ]}

\AtBeginDocument{%
  \providecommand\BibTeX{{%
    \normalfont B\kern-0.5em{\scshape i\kern-0.25em b}\kern-0.8em\TeX}}}

\acmConference[KDD '26]{Proceedings of the 32nd ACM SIGKDD Conference on Knowledge Discovery and Data Mining V.2}{August 09--13, 2026}{Jeju Island, Republic of Korea}
\acmBooktitle{Proceedings of the 32nd ACM SIGKDD Conference on Knowledge Discovery and Data Mining V.2 (KDD '26), August 09--13, 2026, Jeju Island, Republic of Korea}

\begin{document}

\title[Personalizing Marketplace Policies with Competing Objectives and Constrained Experiments]{Personalizing Marketplace Policies with Competing Objectives and Constrained Experiments: Evidence from a Job Marketplace}
\author{Yufei Wu}
\affiliation{%
  \institution{LinkedIn Corporation}
  \city{Sunnyvale}
  \state{CA}
  \country{USA}
}
\email{yufwu@linkedin.com}
\orcid{0009-0003-5951-0099}

\author{Zhen Yan}
\affiliation{%
  \institution{LinkedIn Corporation}
  \city{Mountain View}
  \state{CA}
  \country{USA}
}
\email{zhyan@linkedin.com}
\orcid{0009-0000-7386-537X}

\begin{abstract}

Two-sided marketplaces connect distinct user groups whose interests often conflict---improving outcomes on one side could degrade the other side's experience. To address this challenge, we deploy an integrated framework for personalizing \textbf{free-value thresholds}---a policy governing the scope of complimentary services for job listings---across a two-sided job marketplace connecting millions of employers and job seekers. Our personalized policy delivers \textbf{statistically significant and economically sizable lift in the target metric} while respecting engagement guardrail constraints.

Direct application of standard uplift methods proves insufficient here for two reasons. First, \textbf{cross-side externalities demand multi-objective optimization}: maximizing employer-side metrics risks harming job seeker engagement, with effects varying substantially across job segments. Second, \textbf{marketplace interference necessitates cluster-level randomization}, limiting us to few discrete treatment levels---effectively a form of positivity violation that rules out methods designed for continuous treatments.

We contribute an integrated framework with three components. Our \textbf{ensemble-based hybrid ranking} models target and guardrail metrics separately, cutting guardrail risk by over 10\% for equivalent target gains compared to single-objective approaches. A \textbf{treatment effect extrapolation method} extends our estimates from limited experimental variation to untested policy levels, relying on monotonicity assumptions that we validate empirically. Finally, we present \textbf{production deployment}, where post-launch data confirms both extrapolation accuracy and guardrail compliance.

Our deployed system demonstrates that principled methodology can enable meaningful personalization even when experiments are severely constrained and different objectives compete---common conditions that characterize many real-world marketplaces.

\end{abstract}



\maketitle

\section{Introduction}

\subsection{The Marketplace Policy Personalization Problem}

Two-sided marketplaces face a fundamental tension: optimizing one side's experience often affects the other. Job platforms must balance employer monetization against job seeker engagement; ride-sharing platforms trade driver earnings against rider wait times; e-commerce marketplaces weigh seller outcomes against buyer satisfaction. These tradeoffs are not merely business constraints---they are structural features of platform economics~\citep{edelman2015price,horton2010online}.

\begin{figure}[!htb]
    \centering
    \includegraphics[width=\columnwidth]{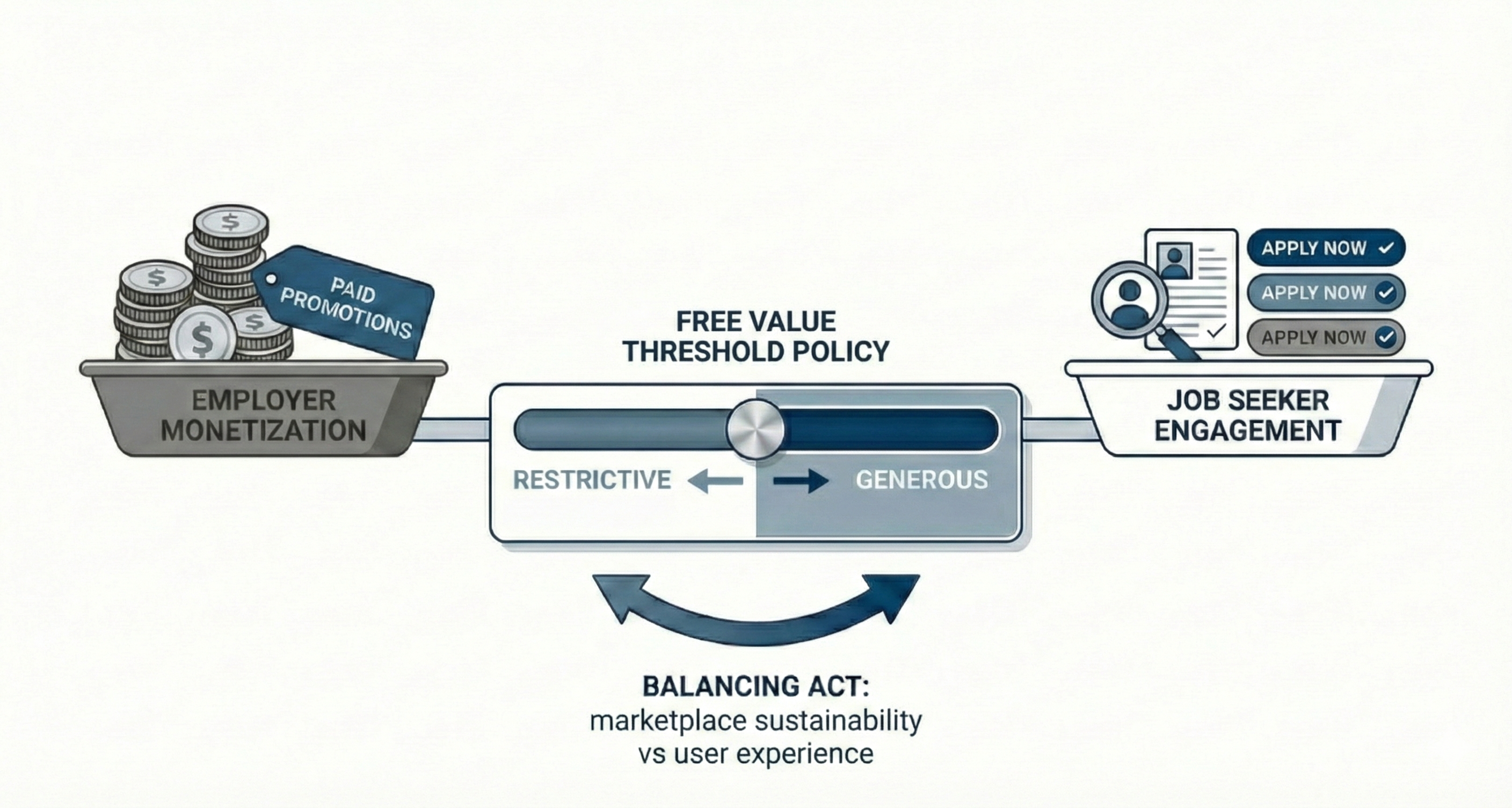}
    \caption{Two-Sided Marketplace Policy Tradeoff.}
    \Description{Diagram illustrating the tradeoff between marketplace sustainability and job seeker engagement under different free-value threshold policies.}
    \label{fig:marketplace-tradeoff}
    \vspace{-1ex}
\end{figure}

We address this challenge in a job marketplace through \textbf{free-value thresholds}---parameters governing the scope of complimentary services for job listings before requiring paid promotion (Figure~\ref{fig:marketplace-tradeoff}). This policy captures a core marketplace tradeoff: driving employer conversions while protecting job seeker engagement. Setting thresholds appropriately is crucial: overly generous thresholds reduce monetization incentives and challenge long-term sustainability; overly restrictive thresholds preserve platform sustainability but may degrade seeker experience.

Most platforms apply \textbf{uniform policies} to all postings. However, job posts vary substantially in seeker engagement, monetization potential, and response patterns, making uniform policies suboptimal. This heterogeneity creates an opportunity to personalize thresholds based on segment-specific responses---improving monetization while protecting job seeker engagement.

\subsection{Technical Challenges}

\textbf{Uplift modeling} has emerged as an established method for policy personalization, estimating conditional average treatment effects (CATE) to target interventions optimally~\citep{devriendt2018literature,kunzel2019metalearners}. However, two interconnected challenges make direct application insufficient in our setting:

\textbf{Challenge 1: Cross-Side Externalities Require Multi-Objective Optimization.}
In two-sided marketplaces, a policy that directly affects one side typically generates indirect effects on the other---a cross-side externality. For example, adjusting free-value thresholds may improve platform sustainability, but risks degrading seeker engagement if access to desirable jobs becomes constrained. Crucially, these effects vary by segment: some jobs convert without engagement loss; others lose engagement without converting. Ultimately, marketplace policies often affect multiple stakeholder groups simultaneously with heterogeneous responses.

Traditional uplift frameworks focus on maximizing a single objective such as conversions~\citep{gubela2020profit,lemmens2020managing,devriendt2020learning}. Recent extensions incorporate budget constraints and ROI thresholds~\citep{google2023fractional,verbeke2022todo,devos2025continuous}, but these formulations layer financial constraints onto a primary objective rather than modeling interacting cross-side effects. Our setting requires explicitly balancing employer-side metrics against job seeker-side engagement, which existing single-objective approaches do not directly capture.

\textbf{Challenge 2: Marketplace Interference Limits Treatment Variation.}
Cross-side externalities create interference effects that complicate experimental design. Recent work~\citep{johari2022experimental,holtz2024cluster} shows that marketplace competition creates interference, leading to biased estimates under individual-level randomization. Cluster-level randomization that minimizes interference addresses this bias but constrains experimental capacity: fewer randomization units limit testing to a few discrete treatment levels rather than the continuous variation that standard dose-response methods require~\citep{hirano2004propensity,kennedy2017nonparametric,athey2019generalized}.

\subsection{Our Approach and Contributions}

We developed an \textbf{integrated framework for policy optimization under positivity violation and cross-side externalities}---two challenges that co-exist in marketplace settings but are typically addressed separately in the literature. Our approach jointly solves: (1) balancing cross-side objectives without contested weights, and (2) optimizing continuous policies when experiments provide only discrete treatment levels.

\textbf{Contribution 1: Ensemble-Based Multi-Objective Hybrid Ranking.}
Single-objective uplift modeling ignores cross-side externalities; standard multi-objective approaches typically require contested weights or navigating complex Pareto frontiers. We contribute a hybrid ranking framework that:
\begin{itemize}
    \item Separately models causal uplift in target and guardrail metrics
    \item Uses ensemble estimates to address spillovers common in marketplaces
    \item Achieves \textbf{over 10\% lower guardrail risk for equivalent target gains} compared to single-objective optimization
\end{itemize}

Our approach identifies a preferred point on the efficient frontier given guardrail constraints, avoiding both contested weight negotiations and the operational complexity of navigating all Pareto-optimal solutions.

\textbf{Contribution 2: Treatment Effect Extrapolation Under Limited Treatment Variation.}
Cluster-randomized marketplace experiments constrain treatment variation to a few discrete levels---insufficient for standard continuous treatment methods. We contribute a principled extrapolation approach that:
\begin{itemize}
    \item Extends CATE estimates from limited experimental levels to untested policy values
    \item Uses validated linear assumptions within bounded ranges
    \item Provides explicit conditions under which extrapolation is appropriate
    \item Demonstrates post-deployment validation of extrapolation accuracy
\end{itemize}

\textbf{Contribution 3: Production Deployment with Post-Launch Validation.}
Industrial uplift applications are often evaluated offline only. We contribute:
\begin{itemize}
    \item Deployment evidence from a system serving millions of job postings and seekers
    \item Post-launch validation showing extrapolated predictions align with observed effects
    \item Practitioner lessons on when ensemble methods and linear extrapolation succeed
\end{itemize}

These contributions form an integrated methodology. The hybrid ranking (Contribution 1) requires reliable CATE estimates at policy levels beyond experimental variation, which the extrapolation method (Contribution 2) provides. Production deployment (Contribution 3) validates that the framework performs as designed under real-world conditions. Together, they constitute a principled approach to a problem class common in multi-sided marketplaces---multi-objective policy optimization under limited treatment variation---that lacks established solutions.

\subsection{Paper Organization}

Section~\ref{sec:related} reviews related work. Section~\ref{sec:methodology} presents our methodology: CATE estimation with ensembles, treatment effect extrapolation, and multi-objective hybrid ranking. Section~\ref{sec:empirical} presents empirical findings from offline analysis: CATE model validation, hybrid ranking results, policy design, and extrapolation robustness. Section~\ref{sec:deployment} covers production deployment: system architecture, post-launch results, extrapolation validation, and practitioner lessons. Section~\ref{sec:limitations} discusses limitations and future work. Section~\ref{sec:conclusion} concludes. 

\section{Related Work}
\label{sec:related}

We position our work at the intersection of three research streams: multi-objective marketplace optimization, heterogeneous treatment effect estimation, and continuous dose-response modeling. To our knowledge, no established framework jointly addresses (A) multi-objective optimization with heterogeneous cross-side effects and (B) policy optimization when treatment variation is limited to few discrete levels---a combination common in marketplace settings.

\subsection{Multi-Objective Optimization in Platforms}

Platforms face inherent multi-objective tradeoffs across stakeholder groups~\citep{edelman2015price,horton2010online}. Recent work addresses these tradeoffs through weighted objectives~\citep{abdollahpouri2020multistakeholder} (requiring contested tradeoff weights) or Pareto optimization~\citep{lin2019pareto} (requiring sophisticated frontier selection).

Most uplift modeling work focuses on single-objective optimization~\citep{gubela2020profit,lemmens2020managing,devriendt2020learning}. Recent extensions incorporate budget constraints~\citep{google2023fractional,verbeke2022todo} and ROI thresholds~\citep{devos2025continuous}, but these layer financial constraints onto a primary objective rather than modeling outcomes across both sides of the marketplace.

\textbf{Gap}: Existing multi-objective frameworks require either contested weight specification or navigating complex Pareto frontiers. Our guardrail-constrained approach provides a simpler mechanism: maximize one objective while explicitly constraining another. This naturally aligns with organizational structures where different teams own different metrics and identifies a single recommended policy on the efficiency frontier.

\subsection{Heterogeneous Treatment Effect Estimation}

Conditional average treatment effect (CATE) estimation has matured rapidly. Meta-learners~\citep{kunzel2019metalearners}, causal forests~\citep{athey2019generalized,wager2018estimation}, and doubly-robust methods~\citep{chernozhukov2018double} provide flexible approaches for binary treatments. The X-learner~\citep{kunzel2019metalearners} provides our base estimator.

Industrial deployments demonstrate CATE's practical value: Uber's CausalML library~\citep{chen2020causalml} enables personalized incentive allocation; applications at Booking.com~\citep{moraes2023uplift} and Kuaishou~\citep{meng2024coarse} show heterogeneous causal learning for resource allocation. However, these applications assume sufficient treatment variation---either from observational data or multi-armed experiments---that our setting lacks.

\textbf{Gap}: Standard CATE estimators require treatment variation across the policy space of interest. Our setting requires extrapolation beyond observed treatment levels.

\subsection{Dose-Response Modeling for Continuous Treatments}

Methods for continuous dose-response estimation have advanced significantly. Generalized propensity score methods~\citep{hirano2004propensity} extend the Rosenbaum-Rubin framework to continuous treatments. The R-learner~\citep{nie2021quasi} provides doubly-robust estimation, while generalized random forests(GRF)~\citep{athey2019generalized} offer nonparametric flexibility. Recent work on dose-response estimation~\citep{kennedy2017nonparametric} handles high-dimensional confounders.

Violation of positivity assumptions poses a fundamental challenge for these methods: ``A challenge for doing causal inference with continuous interventions is that the positivity assumption is typically violated''~\citep{schomaker2024causal}. Practical settings often have sparse support at discrete levels.

\textbf{Gap}: Continuous treatment methods assume dense support across the treatment space---i.e., positivity. Cluster-randomized marketplace experiments necessarily violate this assumption: limited treatment variation is a consequence of addressing interference, not a design flaw. Our extrapolation approach provides a principled solution when positivity cannot be satisfied.

Our work addresses these gaps through an integrated framework. While prior work tackles multi-objective optimization or limited treatment variation in isolation, marketplace settings require solving both simultaneously---the same interference that necessitates cluster randomization with few treatment levels also creates cross-side externalities requiring multi-objective treatment. Our framework addresses this joint problem, which to our knowledge lacks established solutions.

\section{Methodology}
\label{sec:methodology}

\subsection{Problem Formulation}

\textbf{Notation.} For job $j$: features $X_j$ (contextual attributes); policy level $c_j$; target metric outcome $Y_j$; engagement outcome $V_j$.

\textbf{Objective.} Design personalized policy $\pi: X \rightarrow C$ that:
\begin{equation}
\max_{\pi} \sum_{j}\tau_Y(X_j, \pi(X_j)) \quad \text{subject to} \quad \sum_j\tau_V(X_j, \pi(X_j)) \geq -\delta
\label{eq:objective}
\end{equation}
where $\tau_Y(X, c)$ and $\tau_V(X, c)$ are treatment effects on target and guardrail metrics at policy level $c$ versus baseline $c_0$, and $\delta$ is the guardrail threshold.

\textbf{Core empirical challenge.} We observe outcomes only at $c \in \{c_0, c_1\}$ from our experiment, not at other policy-relevant levels.

\subsection{CATE Estimation with Ensemble for Spillovers}

\subsubsection{Base CATE Estimation (X-learner)}
\label{sec:cate-estimation}

We estimate heterogeneous treatment effects using the X-learner~\citep{kunzel2019metalearners}, a flexible meta-learner well-suited for settings with two treatment levels, and benchmark it against alternatives.\footnote{We compare against R-learner and GRF in Appendix~\ref{appendix:cate-comparison}: X-learner and R-learner agree on both target ($\rho = 0.74$) and guardrail ($\rho = 0.89$), cross-validating the X-learner choice; GRF agrees on guardrail ($\rho = 0.76$--$0.78$) but diverges on target ($\rho = 0.35$--$0.36$), consistent with GRF's local kernel-weighted forest estimates being sensitive to per-leaf positive sparsity under rare-positive signal.} While our randomized experiment ensures balanced treatment assignment, our target metric (paid conversions) exhibits class imbalance. Therefore, we take a two-stage approach---modeling outcomes separately for treated and control groups, then imputing counterfactuals---which provides more stable estimates than single-model approaches.

\textbf{Step 1: Outcome Models.} Fit models separately for control and treatment:
\begin{equation}
\hat{\mu}_0(X) = \mathbb{E}[Y | X, W=0], \quad \hat{\mu}_1(X) = \mathbb{E}[Y | X, W=1]
\end{equation}

We use gradient boosted trees (XGBoost) with:
\begin{itemize}
    \item \textbf{Train-validation-holdout splits at the employer level} to prevent leakage and reduce overfitting
    \item \textbf{Target encoding for high-cardinality contextual features}, pooling signals across sparse segments while avoiding dimensionality explosion~\citep{micci2001preprocessing}
    \item \textbf{Two-stage modeling for class imbalance}: For the target metric, we employ a two-stage approach: (1)~predict the binary conversion outcome using balanced class weights; (2)~conditional on conversion, predict continuous outcome using only positive examples. This decomposition improves calibration for rare events while capturing response heterogeneity among the positive class.
\end{itemize}

\textbf{Step 2: Cross-Imputation of Treatment Effects.} Compute pseudo-treatment effects:
\begin{equation}
\tilde{\tau}_1(X_i) = Y_i - \hat{\mu}_0(X_i) \text{ for treated}, \quad \tilde{\tau}_0(X_i) = \hat{\mu}_1(X_i) - Y_i \text{ for control}
\end{equation}

\textbf{Step 3: Predict CATE.} Train models $\hat{\tau}_1(X)$ and $\hat{\tau}_0(X)$ on the imputed effects, then combine them via propensity-weighted average. Given random assignment with equal propensities:
\begin{equation}
\hat{\tau}(X) = 0.5 \cdot \hat{\tau}_0(X) + 0.5 \cdot \hat{\tau}_1(X)
\end{equation}

\subsubsection{Ensemble for Cross-Job Spillovers}

The policy must be implemented at the job-segment level to ensure consistency across employers posting similar jobs. However, employers posting multiple jobs create spillover risk: changing policy for one job may shift behavior across their postings. This creates an aggregation challenge: job-level CATE captures direct effects but ignores spillovers across jobs within the same employer; employer-level CATE captures net effects but cannot provide policies varying only by job segments.

To address this, we ensemble CATE estimates on job-level outcomes and employer-level average outcomes:
\begin{itemize}
    \item \textbf{Job-level CATE} $\hat{\tau}_{\text{job}}(X)$: Predicts job-level outcome. Most appropriate for single-job employers.
    \item \textbf{Employer-level CATE} $\hat{\tau}_{\text{employer}}(X)$: Predicts average outcome across all postings of the same employer. Most appropriate for multi-job employers.
\end{itemize}

Final estimate combines the two CATE estimates, weighted by segment composition. $w_{\text{single}}$ and $w_{\text{multi}}$ denote the share of jobs in a segment posted by employers with single vs.\ multiple jobs:
\begin{equation}
\hat{\tau}_{\text{final}}(X) = w_{\text{single}} \hat{\tau}_{\text{job}}(X) + w_{\text{multi}} \hat{\tau}_{\text{employer}}(X)
\end{equation}

In segments dominated by single-job employers, job-level CATE dominates; in segments with many multi-job employers, employer-level CATE receives higher weight. This approach is analogous to hierarchical modeling strategies used in marketplace experimentation~\citep{johari2022experimental} and multi-level CATE estimation~\citep{kunzel2019metalearners}, adapted to our constraint that targeting occurs at the job-segment level while outcomes aggregate at the employer level.

\subsection{Treatment Effect Extrapolation}
\label{sec:extrapolation}

Our experiment allows estimating CATE $\hat{\tau}(X, c_1)$ (effect at treatment vs.\ baseline). We need estimates at other policy-relevant levels which were not tested.

We extrapolate treatment effects linearly from the observed experimental contrast:
\begin{equation}
\hat{\tau}(X, c) = \hat{\tau}(X, c_1) \cdot \frac{c - c_0}{c_1 - c_0}
\label{eq:extrapolation}
\end{equation}

\textbf{Linear extrapolation as an identification constraint.} We emphasize that linearity here is not a modeling preference but a consequence of the experimental design: with only two observed treatment levels $\{c_0, c_1\}$, a linear functional form is the highest-degree dose-response that the data identify. Fitting any higher-order polynomial in $c$ would require either three or more treatment levels --- which our existing experiment data did not provide --- or additional functional-form assumptions, which we do not impose; fitting nonparametric continuous dose-response would in addition require dense support across the treatment space~\citep{hirano2004propensity,kennedy2017nonparametric,schomaker2024causal}. We therefore treat Equation~\ref{eq:extrapolation} as the maximal model that the design identifies, not as a claim that the underlying dose-response is globally linear. We are not asserting that dense treatment support is unnecessary --- on the contrary, dense support remains essential for capturing non-linearities (e.g., saturation effects) and for providing cardinal magnitude estimates needed for finer-grained, continuous personalization policies. Among the family of two-parameter dose-response forms compatible with our experimental variation, we choose the linear-in-$c$ form for its parsimony and interpretability; alternative monotone transformations (log-linear, exponential) fit our two observed means equally well but are not separately identifiable from them. The validity of the resulting policy thus rests on two complementary supports: (i)~the policy assigns tiered levels by \emph{ordinal rank} of estimated CATE rather than by cardinal magnitude (Section~\ref{sec:hybrid-ranking}), which is preserved under any monotonic transformation of the dose-response; and (ii)~the deployed policy range stays within $\sim$50\% of the observed experimental variation~$(c_1 - c_0)$, the regime within which our offline sensitivity analysis (Section~\ref{sec:extrapolation-robustness}) and post-deployment validation (Section~\ref{sec:extrapolation-validation}) confirm that policy recommendations are robust to plausible deviations from linearity.

We validate this approach through offline sensitivity analysis (Section~\ref{sec:extrapolation-robustness}) and post-deployment comparison of extrapolated predictions with observed outcomes (Section~\ref{sec:extrapolation-validation}).

\textbf{Toward continuous dose-response: complementary paths.} The identification constraint above motivates two complementary research directions for relaxing the two-level limitation. First, richer experimental designs that add discrete treatment levels --- e.g., a multi-arm holdout with three or more levels --- would extend the identifiable regime and, when treatment support is sufficiently dense, allow flexible continuous dose-response estimation via the R-learner~\citep{nie2021quasi} or generalized random forests~\citep{athey2019generalized}. Second, subsequent iterations of our system can integrate \emph{in-model exploration} --- contextual bandits and online policy-learning schemes~\citep{slivkins2019introduction, amani2019linear} that adaptively sample policy levels as a function of context $X$ while respecting safety constraints during exploration. Together, these would let the framework deliver cardinal-magnitude policy recommendations rather than the ordinal-rank policy classes used in the present deployment. Designing in-model exploration that respects cross-side guardrails and cluster-level interference is a separate problem we leave for future work (Section~\ref{sec:future-directions}).

\subsection{Multi-Objective Hybrid Ranking}
\label{sec:hybrid-ranking}

Job segments likely respond heterogeneously to policy changes: some may monetize without engagement loss, while others may lose engagement without monetizing. Importantly, target uplift and guardrail risk need not be correlated---optimizing solely for the target metric could overlook segments with disproportionate engagement impact. This motivates a \textbf{hybrid ranking algorithm} that optimizes for target metric while explicitly controlling guardrail risk.

\subsubsection{Hybrid Ranking Algorithm}

\textit{Step 1.} Estimate separate ensemble CATEs for target metric ($\hat{\tau}_Y(X)$) and engagement ($\hat{\tau}_V(X)$).

\textit{Step 2.} Create hybrid ranking:
\begin{itemize}
    \item \textbf{Top half}: High to medium $\hat{\tau}_Y(X)$ (target metric uplift), ranked in descending order
    \item \textbf{Bottom half}: Among jobs with low $\hat{\tau}_Y(X)$, rank by $\hat{\tau}_V(X)$ (engagement impact) from least to most negative
\end{itemize}

\textit{Step 3.} Assign tiered policies based on ranking (Table~\ref{tab:tiers}).

\begin{table}[!htb]
\caption{Policy tier assignment based on hybrid ranking.}
\label{tab:tiers}
\centering
\footnotesize
\begin{tabular}{@{}lllll@{}}
\toprule
Group & \makecell{Target\\Lift} & \makecell{Guardrail\\Risk} & Policy & Rationale \\
\midrule
A & High & Low & Most restrictive & Maximize target \\
B & Medium & Low/Med & Moderately restrictive & Balance target vs guardrail \\
C & Low & Low & Baseline & No gain/risk \\
D & Low & High & Relaxed & Protect engagement \\
\bottomrule
\end{tabular}
\end{table}

\textit{Step 4.} Verify guardrail constraint: $\sum_{j} \hat{\tau}_V(X_j, \text{policy}_j) \geq -\delta$. If violated, relax policies for lower-ranked segments iteratively until constraint is satisfied.

\subsubsection{Why Hybrid Ranking Over Alternatives}

Alternative approaches include:
\begin{itemize}
    \item \textbf{Weighted composite objective}: Combine target and guardrail into a single weighted objective $U = w_Y \tau_Y + w_V \tau_V$. However, this requires stakeholders to agree on weights a priori, which proved infeasible across teams representing different marketplace sides.
    \item \textbf{Pareto optimization with frontier selection}: Identify all non-dominated policies, then let stakeholders choose. While theoretically comprehensive, presenting multiple Pareto-optimal solutions proved difficult to operationalize---stakeholders preferred a single recommended policy, not a frontier to navigate.
\end{itemize}

Our hybrid ranking identifies a preferred policy on the efficiency frontier through explicit constraints rather than weights. The guardrail threshold $\delta$ implicitly determines a point on the Pareto frontier, but does so through interpretable business constraints rather than contested preference weights. Additionally, ranking by impact (ordinal) is more robust to noise than optimizing weighted objectives (cardinal).\footnote{\label{fn:tradeoff}We quantify this under $100$ noise realizations ($\sigma = $ bootstrapped SE per unit): compared to exact Lagrangian optimization, hybrid ranking captures approximately $8\%$ less gain in the target metric while honoring the guardrail constraint, but more than halves the policy-flip rate under noise --- reflecting hybrid's invariance to $\hat{\tau}_V$ noise by construction.}

\subsubsection{Extension to More Than Two Objectives}
\label{sec:multi-objective-extension}

The hybrid ranking framework extends naturally beyond a single target and a
single guardrail. We sketch two complementary generalizations below.

\textit{Approach A: Dimensional aggregation.} When multiple target
metrics $Y_1, \ldots, Y_m$ are owned by a single team and a composite
metric is tractable, define
$\tilde{Y} = \sum_{k=1}^{m} w^Y_k \, Y_k$ with relevant weights
$\{w^Y_k\}$, and similarly
$\tilde{V} = \sum_{k=1}^{\ell} w^V_k \, V_k$ for the guardrail side.
This reduces the dimensionality of the problem, allowing the
hybrid ranking approach to apply unchanged to the composite CATE
estimates $(\hat{\tau}_{\tilde{Y}}, \hat{\tau}_{\tilde{V}})$.

\textit{Approach B: Generalized constrained optimization.} When guardrail
metrics cannot be sensibly aggregated---for instance, when each guardrail
$V_k$ has its own threshold $\delta_k$ set by a different
stakeholder---we could either apply hybrid ranking based on the
most-binding guardrail (sorting low-target-lift jobs by
$\min_k \tau_{V_k}(X_j,\cdot)/\delta_k$), or solve a generalized
constrained optimization problem:
\[
\max_{\pi} \sum_{j} \tau_Y(X_j, \pi(X_j))
\;\;\text{s.t.}\;\;
\sum_{j} \tau_{V_k}(X_j, \pi(X_j)) \geq -\delta_k,
\;\; k = 1, \ldots, \ell.
\]
The choice depends on the optimality-stability tradeoff reported
empirically in Footnote~\ref{fn:tradeoff}.

\section{Empirical Findings}
\label{sec:empirical}

\subsection{Data}

\textbf{Data source}: Cluster-randomized experiment
\begin{itemize}
    \item \textbf{Sample}: Jobs posted within the multi-week experiment
    \item \textbf{Arms}: Control (baseline policy) vs.\ Treatment (alternative policy level)
    \item \textbf{Randomization}: Cluster-level to minimize marketplace interference
\end{itemize}

Exact sample sizes are proprietary, but the experiment includes sufficient statistical power for heterogeneous effect estimation. Treatment assignment balance was confirmed via sample ratio mismatch (SRM) testing and pre-experiment A/A validation.

\textbf{Outcomes}:
\begin{itemize}
    \item Target metrics: Employer conversion event (binary), monetization outcome (continuous)
    \item Guardrail: Job seeker engagement metrics
\end{itemize}

\textbf{Features}: Approximately 10 contextual attributes

\textbf{Data splits} (employer-level to prevent leakage): Training 60\% | Validation 20\% | Holdout 20\%

\subsection{CATE Model Validation}

We first validate that our CATE models produce reliable predictions before using them for policy optimization.

\textbf{Calibration.} Using holdout data, predicted CATE correlates strongly with observed treatment effects when grouping jobs by predicted score: 0.95 (target) and 0.86 (guardrail) at the decile level. Correlations remain robust at finer granularity (20 buckets: 0.79 target, 0.75 guardrail), indicating stable predictive accuracy across aggregation levels. This confirms that model predictions reliably reflect true heterogeneous responses in out-of-sample evaluation.

\begin{figure}[!htb]
    \centering
    \captionsetup[subfigure]{font=footnotesize}
    \begin{subfigure}[b]{0.48\columnwidth}
        \includegraphics[width=\textwidth]{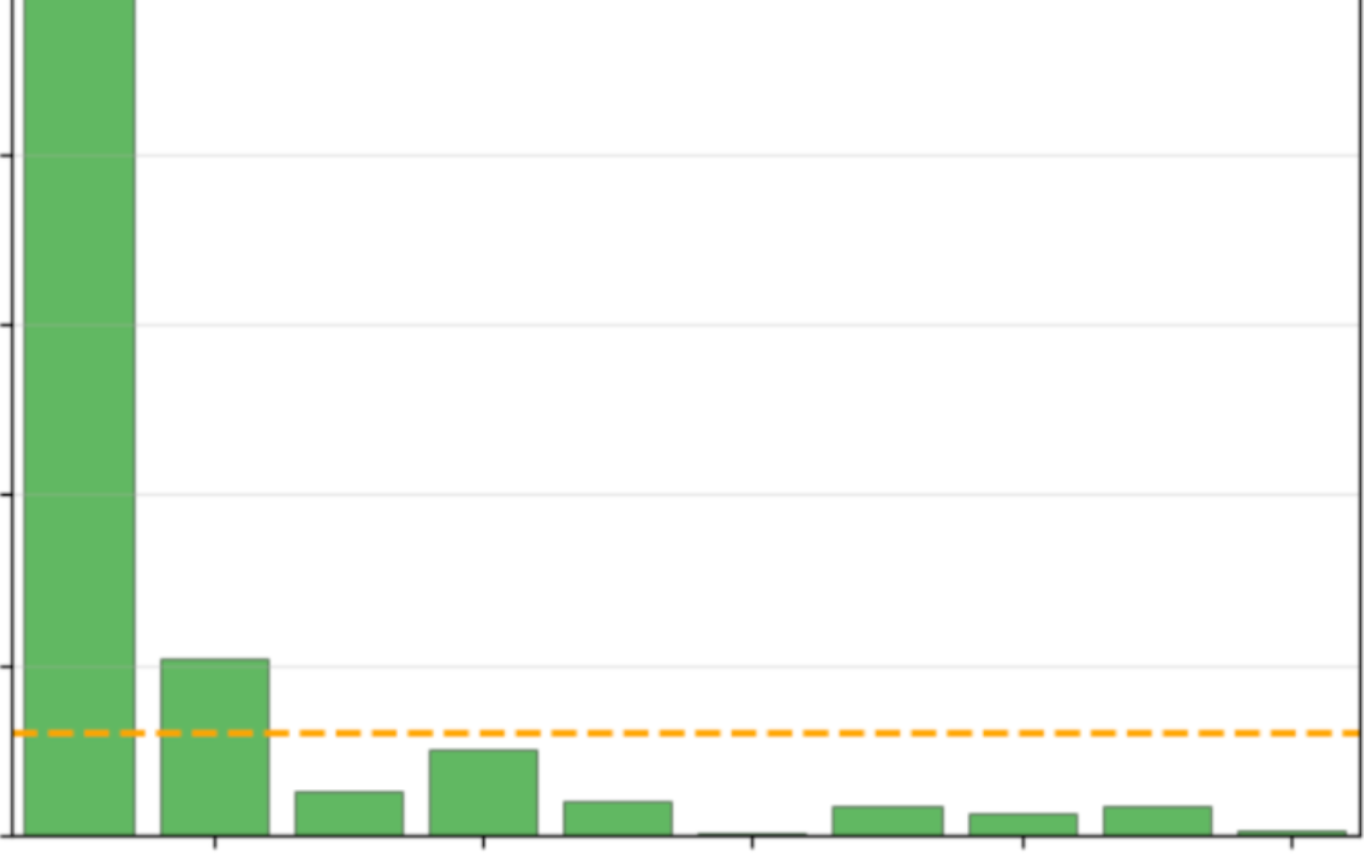}
        \caption{Target Effect by Target CATE Ranking}
    \end{subfigure}
    \hfill
    \begin{subfigure}[b]{0.48\columnwidth}
        \includegraphics[width=\textwidth]{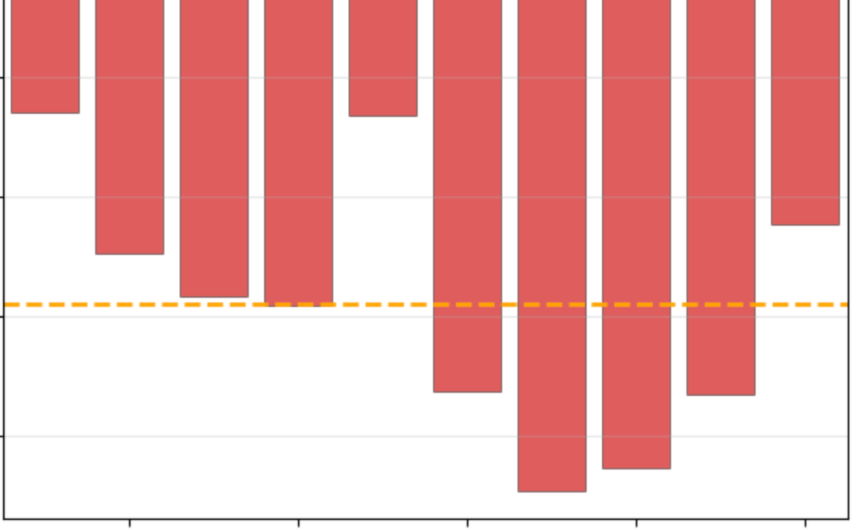}
        \caption{Guardrail Effect by Target CATE Ranking}
    \end{subfigure}
    \caption{CATE Model Evaluation using Holdout Data.}
    \Description{Two bar charts showing observed treatment effects by predicted CATE decile. Left panel shows target effects declining monotonically from top to bottom decile. Right panel shows guardrail effects with non-monotonic pattern across deciles.}
    \label{fig:cate-validation}
    \vspace{-1ex}
    {\small\textit{Note:} Orange dashed line indicates ATE benchmark.}
\end{figure}

\textbf{Ranking quality.} Figure~\ref{fig:cate-validation} shows observed treatment effects by predicted target CATE decile using holdout data. The left panel demonstrates that ranking by predicted CATE successfully identifies high-value segments: the top decile exhibits effects well above ATE, with a generally monotonic decline through middle deciles to near-zero effects in the bottom deciles. The Qini coefficient---measuring ranking quality where 0 indicates random targeting and 1 indicates perfect ranking---reaches \textbf{0.80}, indicating our model captures 80\% of the maximum achievable uplift gain in the target metric.

\textbf{The case for multi-objective optimization.} The right panel of Figure~\ref{fig:cate-validation} reveals why single-objective optimization is insufficient: guardrail effects are \emph{non-monotonic} with target CATE ranking. Risk varies widely across deciles with no consistent pattern. Notably, segments with similar low target effects can have dramatically different guardrail impacts---some well below ATE, others well above. This heterogeneity means optimizing for target alone would apply identical policies to segments with vastly different engagement risks, missing the opportunity to differentiate treatment where guardrail protection matters most. This motivates our hybrid ranking approach.

\subsection{Hybrid Ranking Effectiveness}
\label{sec:hybrid-ranking-results}

\begin{figure}[!htb]
    \centering
    \captionsetup[subfigure]{font=footnotesize}
    \begin{subfigure}[b]{0.48\columnwidth}
        \includegraphics[width=\textwidth]{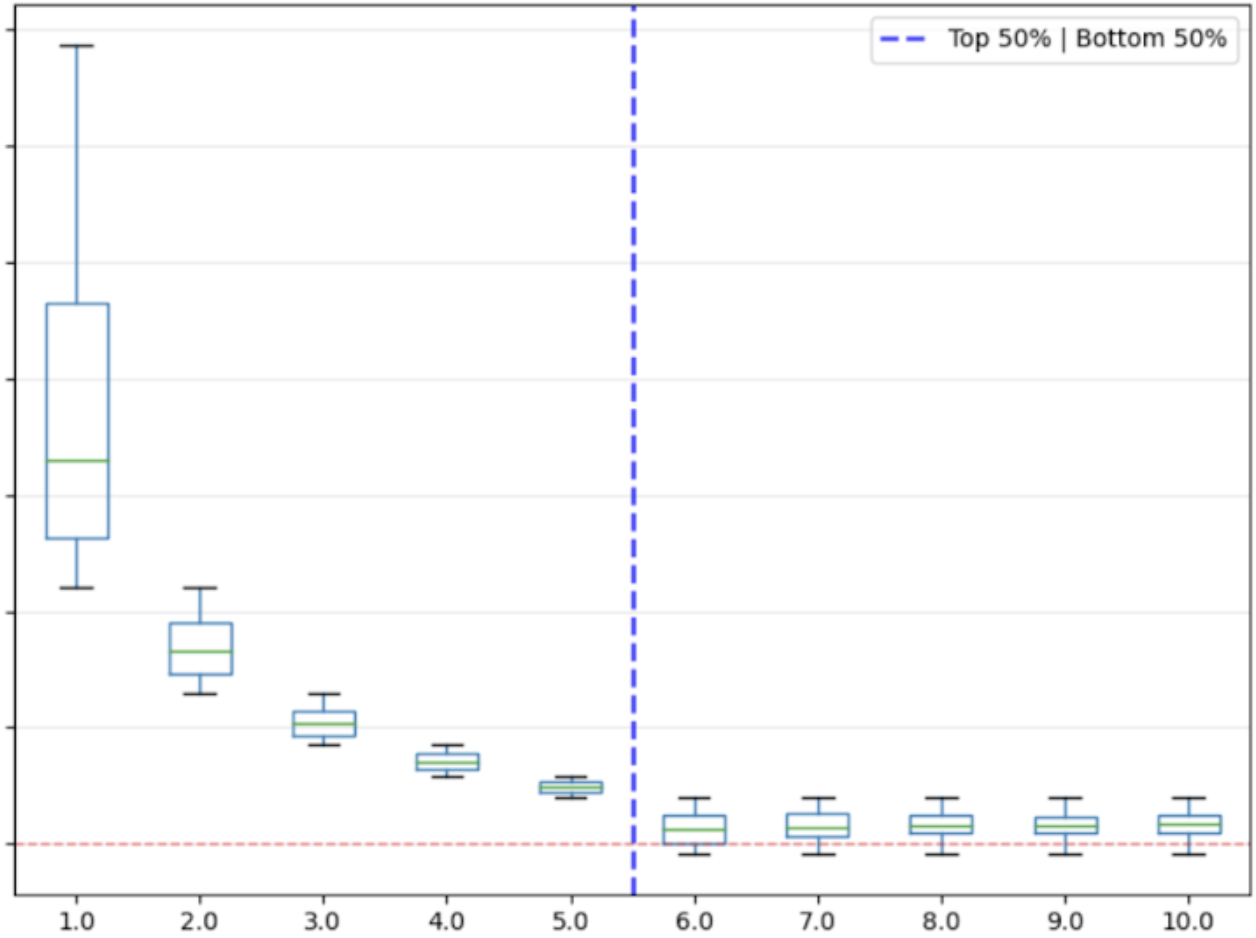}
        \caption{Target CATE by Hybrid Rank}
    \end{subfigure}
    \hfill
    \begin{subfigure}[b]{0.48\columnwidth}
        \includegraphics[width=\textwidth]{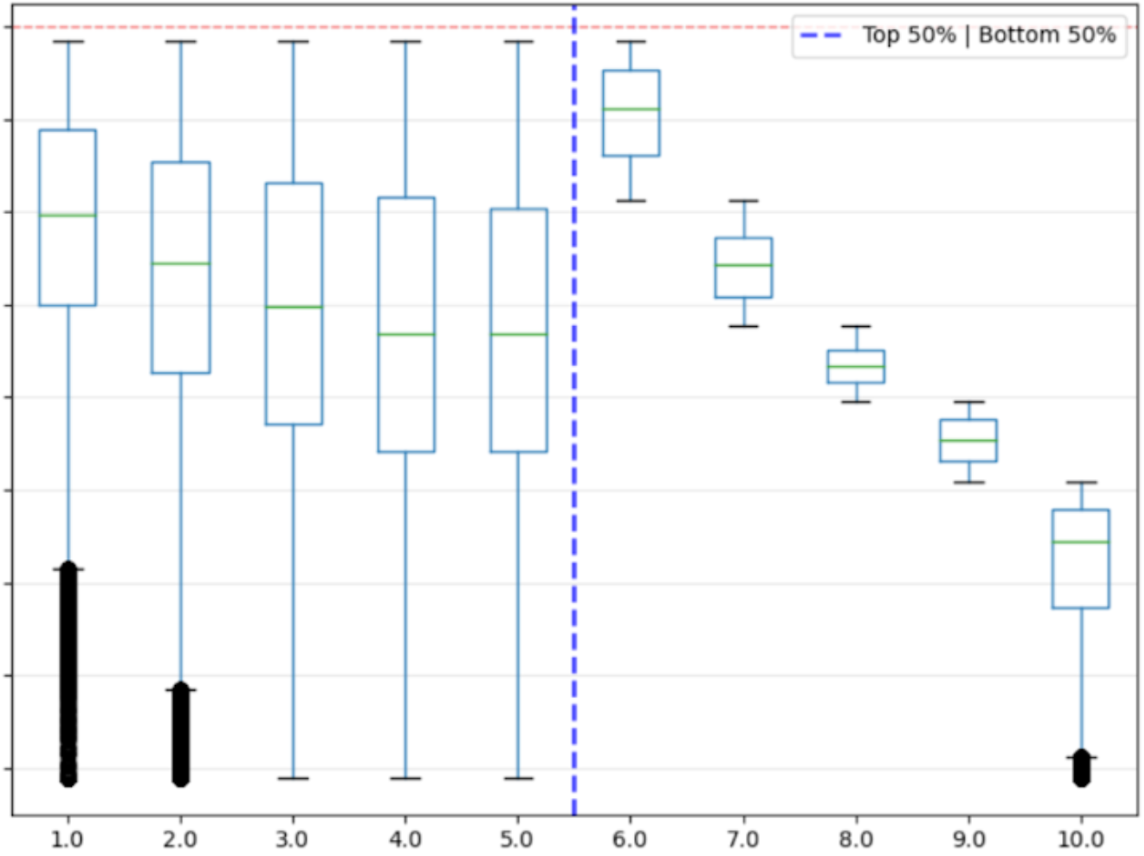}
        \caption{Guardrail CATE by Hybrid Rank}
    \end{subfigure}
    \caption{Target and Guardrail CATE by Hybrid Ranking.}
    \Description{Two bar charts showing CATE values by hybrid ranking position. Left panel shows target CATE high for top-ranked jobs and declining. Right panel shows guardrail CATE sorted to minimize negative engagement impact in the bottom half.}
    \label{fig:hybrid-ranking}
    \vspace{-1ex}
    {\small\textit{Note:} Orange dashed line indicates ATE benchmark.}
\end{figure}

Figure~\ref{fig:hybrid-ranking} illustrates how hybrid ranking addresses the heterogeneity revealed above. The key insight is that the \textbf{top 50\%} of jobs by target CATE offer meaningful monetization opportunities with manageable engagement risk, while the \textbf{bottom 50\%} contribute minimal target uplift but vary substantially in guardrail risk.

\textbf{Connecting to policy tiers.}
\begin{itemize}
    \item \textbf{Left panel (target CATE)}: Top 50\% shows high to medium uplift, corresponding to Groups A and B (Table~\ref{tab:tiers})---these receive restrictive policies to capture upsell upside.\footnote{These findings are robust to CATE estimation noise: observed treatment effect by hybrid ranking is consistent with this ordering across 100 bootstrapped refits; confidence intervals are reported in Appendix~\ref{appendix:bootstrap-stability}, Figure~\ref{fig:bootstrap-hybrid-ate}.}
    \item \textbf{Right panel (guardrail CATE)}: Bottom 50\% has near-zero target uplift but critical guardrail differentiation---risk ranges from close to zero to highly negative. These correspond to Groups C and D---Group C (low risk) keeps baseline; Group D (high risk) receives relaxed policy to protect engagement.
\end{itemize}

\textbf{Quantifying the benefit.} Table~\ref{tab:qini} compares ranking quality using the Qini coefficient. Hybrid ranking has \textbf{little impact on target Qini} (0.83 vs.\ 0.80) while \textbf{nearly doubling guardrail Qini} (0.21 vs.\ 0.11, +91\%). The low guardrail Qini for target-only optimization (0.11, barely above random) confirms that optimizing purely for target effectively ignores guardrail considerations. Our hybrid-ranking approach addresses that limitation. Under the recommended policy, our hybrid ranking reduces guardrail risk by over 10\% compared to target-only optimization while maintaining the target gains.

\begin{table}[!htb]
\caption{Ranking Method Comparison (Qini Coefficient)}
\label{tab:qini}
\centering
\begin{tabular}{@{}lcc@{}}
\toprule
Ranking Method & Target Qini & Guardrail Qini \\
\midrule
Target-Only & 0.80 & 0.11 \\
\textbf{Hybrid (Ours)} & \textbf{0.83} & \textbf{0.21} \\
\midrule
Improvement & +3.5\% & +91\% \\
\bottomrule
\end{tabular}
\end{table}

\textbf{Why the asymmetric benefit?} Reordering the bottom 50\% by guardrail risk has little impact on target ranking quality---these jobs have near-zero target uplift regardless of ordering. But for guardrail protection, this reordering is crucial: it ensures jobs where engagement may be constrained receive relaxed policies rather than being treated identically to low-risk jobs. The hybrid approach maintains target optimization while substantially improving guardrail protection by identifying at-risk segments that single-objective optimization would overlook.

\subsection{Policy Design}
\label{sec:policy-design}

\begin{figure}[!htb]
    \centering
    \includegraphics[width=0.9\columnwidth]{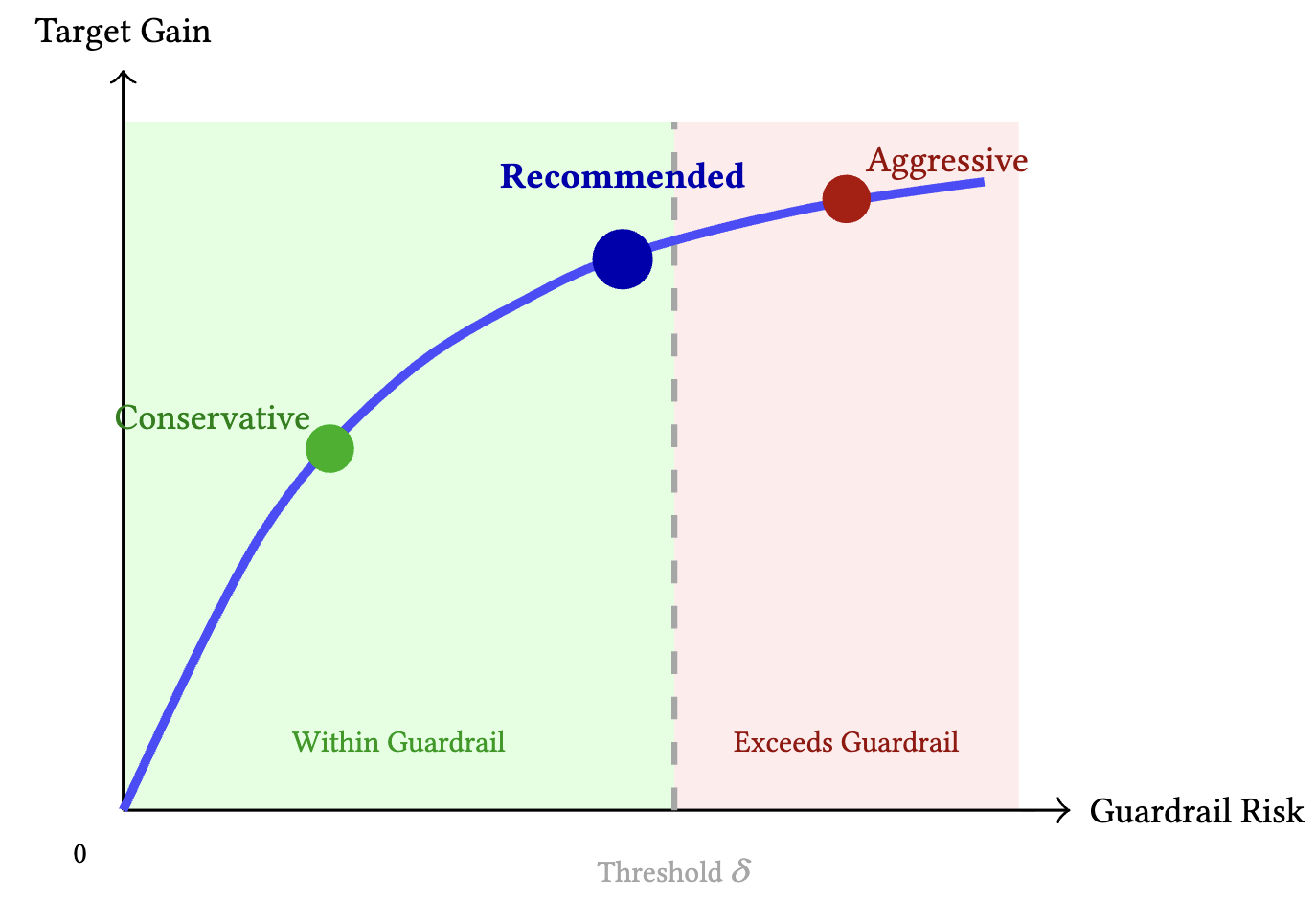}
    \caption{Policy Options on Pareto Efficiency Frontier.}
    \Description{Scatter plot showing three policy options along the Pareto frontier of target gain versus guardrail risk, with a vertical dashed line indicating the guardrail threshold. The recommended policy sits at the threshold boundary.}
    \label{fig:pareto-frontier}
\end{figure}

Figure~\ref{fig:pareto-frontier} visualizes candidate policies along the efficient frontier, with the guardrail threshold shown as a vertical dashed line. The threshold $\delta$ was determined by business stakeholders based on acceptable engagement risk---our framework takes this constraint as input rather than optimizing it. We evaluated three configurations: a \emph{conservative} policy with lower target gain but minimal guardrail risk, our \emph{recommended} policy that maximizes target gain while remaining at the guardrail threshold, and an \emph{aggressive} policy with higher target gain but unacceptable guardrail risk (rejected for exceeding the threshold). While we visualize these options to illustrate tradeoffs, the hybrid ranking algorithm directly produces the recommended policy without requiring ex-post frontier navigation.

The recommended policy strikes a balance facilitating stakeholder alignment:
\begin{itemize}
    \item \textbf{Target metric lift}: Captures majority of potential gain
    \item \textbf{Engagement impact}: Within guardrail threshold
\end{itemize}

This policy represents a point on the efficient frontier identified through explicit guardrail constraints rather than weight negotiation.

\subsection{Extrapolation Robustness}
\label{sec:extrapolation-robustness}

How robust are policy recommendations to the linear extrapolation assumption?

We tested \textbf{sensitivity of policy impact} under alternative curvature assumptions: 20\% concave (diminishing returns) and 20\% convex (increasing returns), corresponding to plausible deviations from linearity within our bounded extrapolation range.

\begin{table}[!htb]
\caption{Extrapolation Sensitivity Analysis}
\label{tab:sensitivity}
\centering
\begin{tabular}{@{}ll@{}}
\toprule
Assumption & Est.\ Impact \\
\midrule
Linear (baseline) & Baseline \\
20\% concave & $+8$\% \\
20\% convex & $-9$\% \\
\bottomrule
\end{tabular}
\end{table}

Table~\ref{tab:sensitivity} summarizes the results. Reassuringly, policy impact varies by only 8--9\% under alternative curvature assumptions. Since hybrid ranking depends on ordinal rankings rather than cardinal magnitudes, this moderate variation does not affect policy recommendations---relative comparisons across policies remain robust.

\section{Production Deployment and Results}
\label{sec:deployment}

We established model quality through offline validation in Section~\ref{sec:empirical}. This section presents evidence from production deployment testing real-world impact.

\subsection{System Architecture}

\textbf{Deployment timeline}:
\begin{itemize}
    \item Phase 1: Multi-week online A/B test with recommended policy
    \item Phase 2: Readout and ramp decision
    \item Phase 3: Ramp to 100\% of eligible jobs
\end{itemize}

\begin{figure}[!htb]
    \centering
    \includegraphics[width=\columnwidth]{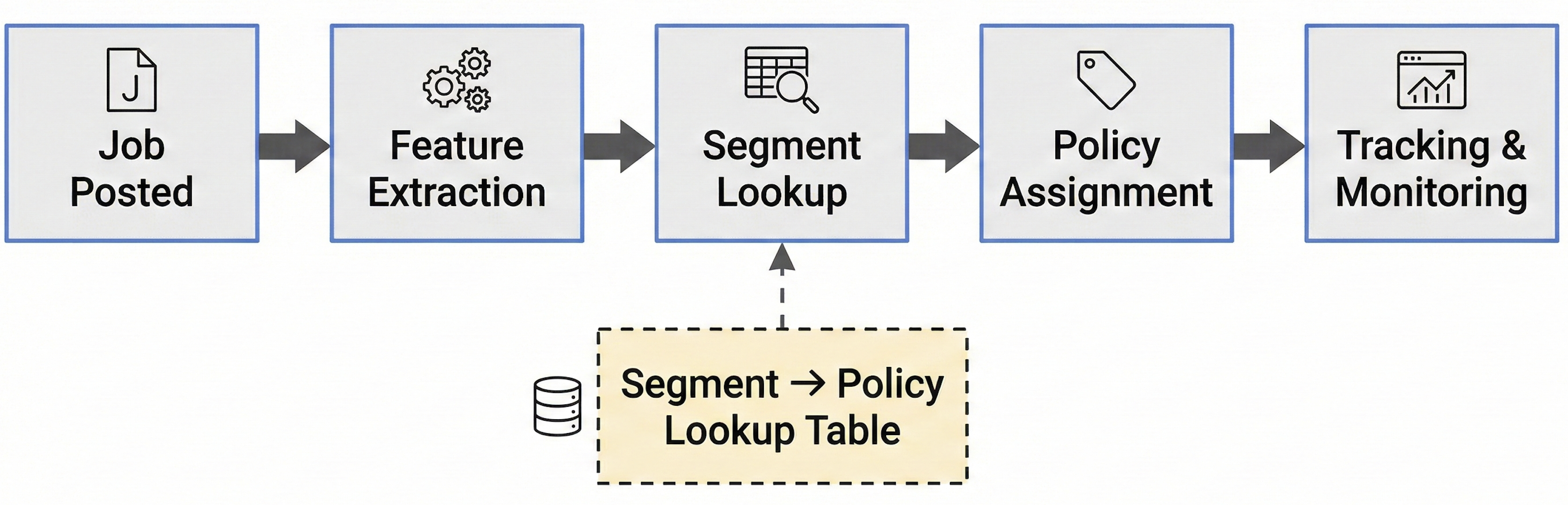}
    \caption{Production System Architecture.}
    \Description{Flow diagram showing the production system: offline CATE model training produces segment-level policy lookup tables, which are served via low-latency lookup with fallback to default policy.}
    \label{fig:architecture}
\end{figure}

\textbf{Production implementation}:
We deployed segment-based policy lookup tables rather than real-time model inference. Segment-level targeting is required to ensure consistent treatment for similar jobs---varying policy at finer granularity would create inconsistent employer experiences. Given this constraint, lookup tables provide additional operational benefits: sub-second latency, no model-serving infrastructure, and easier auditability and debugging. Figure~\ref{fig:architecture} illustrates the system architecture.

\textbf{Engineering considerations}:
\begin{enumerate}
    \item \textbf{Latency}: Sub-second lookup for real-time decisions
    \item \textbf{Infrastructure reuse}: Leveraged existing feature pipeline to reduce engineering cost
    \item \textbf{Feature consistency and fallback}: Ensure online features match offline training; default to baseline policy on lookup failure
    \item \textbf{Gradual validation}: Use test jobs and small ramps to catch implementation bugs before full deployment
    \item \textbf{Stratified monitoring}: Track policy distribution and metrics overall and by tier to detect issues early
\end{enumerate}

\subsection{Post-Launch Results}

Table~\ref{tab:results} summarizes post-launch performance from the online experiment.

\begin{table}[!htb]
\caption{Post-Launch Results Summary}
\label{tab:results}
\centering
\footnotesize
\begin{tabular}{@{}llll@{}}
\toprule
Metric & Predicted & Observed & Status \\
\midrule
Target lift & $\theta$\% & $\sim$0.9$\theta$\% & Stat.\ sig., within 10\% of prediction \\
Guardrail & $\geq -\delta$ & $> -\delta$ & Within threshold \checkmark \\
\bottomrule
\end{tabular}
\end{table}

\textbf{Calibration}: Observed target lift was statistically significant and sizable, within 10\% of the offline prediction---representing acceptable calibration error for production systems. This is consistent with the sensitivity analysis in Section~\ref{sec:extrapolation-robustness}, which showed policy impact could vary by 8--9\% under alternative curvature assumptions.

\textbf{Guardrail Performance}: All engagement metrics remained within guardrail thresholds, validating the hybrid ranking approach. The separate modeling of target and engagement metrics enabled maximizing one while protecting the other.

\subsection{Extrapolation Validation}
\label{sec:extrapolation-validation}

A critical question is whether linear extrapolation from limited experimental variation produces accurate predictions at untested policy levels. Our gradual ramp deployment provided an opportunity to validate this assumption.

\textbf{Validation approach}: During the ramp phase, we deployed intermediate policy levels not included in the original experiment. This created natural test data: we compared extrapolated CATE predictions (generated from the two-level experiment) against observed treatment effects at these intermediate levels.

\textbf{Key findings}:
\begin{enumerate}
    \item \textbf{Strong correlation}: Observed treatment effects correlated highly with linear predictions across policy levels ($r = 0.85$), indicating that the linear assumption captures the dominant structure of the dose-response relationship.
    \item \textbf{Segment-level robustness}: Observed segment rankings generally matched predicted rankings. Even where absolute magnitudes differed, ordinal rankings---which drive policy assignment---remained stable.
    \item \textbf{Tail behavior}: Larger deviations appeared in segments with extreme predicted uplift. This is expected: sparse data in tails increases prediction uncertainty, and behavioral responses may be nonlinear at extremes.
    \item \textbf{Approximate monotonicity}: Treatment effects increased approximately monotonically with policy restrictiveness, with minor deviations attributable to sampling noise rather than systematic nonlinearity.
\end{enumerate}

\textbf{Implications for practitioners}: Linear extrapolation proves reliable when:
\begin{itemize}
    \item The extrapolation range is limited (we stayed within $\sim$50\% of observed treatment variation)
    \item Policy decisions depend on ordinal rankings rather than precise effect magnitudes
    \item Validation data from gradual ramps can verify predictions before full deployment
\end{itemize}

This post-deployment validation confirms the validity conditions identified in Section~\ref{sec:extrapolation} and supports the broader applicability of extrapolation approaches for limited-variation settings.

\section{Limitations and Future Work}
\label{sec:limitations}

\subsection{Limitations}

\textbf{Extrapolation assumptions and range.} Our linear dose-response assumption is validated within the deployment range ($\sim$50\% of observed experimental variation), but likely breaks down at policy extremes. Very low thresholds may face behavioral threshold effects; very high thresholds may encounter diminishing returns. As discussed in Section~\ref{sec:extrapolation}, this linearity is an identification constraint imposed by the two discrete treatment levels in our cluster-randomized experiment, not a modeling choice; relaxing it requires more treatment levels or denser treatment support, either via richer experimentation or in-model exploration that respects cross-side guardrails (see Section~\ref{sec:future-directions}).

\textbf{External validity.} Our results reflect the specific structure of our job marketplace. Applicability to other platforms depends on whether similar conditions hold: heterogeneous treatment effects, approximately linear dose-response within relevant ranges, and cross-side externalities that make multi-objective optimization valuable.

\textbf{Long-term impact.} The deployment evidence in Sections~\ref{sec:deployment} and~\ref{sec:extrapolation-validation} characterizes short-term treatment effects. Several questions of practical importance unfold on longer timescales: whether engagement guardrails monitored at launch persist or attenuate as marketplace participants adapt, and how the policy performs against longer-term downstream outcomes absent from the original experiment's measurement window. We discuss long-term holdout in Section~\ref{sec:future-directions}.

\subsection{Future Directions}
\label{sec:future-directions}

\textbf{Richer experiments and continuous treatment methods.} Multi-armed experiments with diverse treatment levels would enable direct estimation of dose-response curves without extrapolation, allowing testing for nonlinearities. With denser treatment variation, methods like the R-learner~\citep{nie2021quasi} or generalized random forests~\citep{athey2019generalized} could capture nonlinear effects our linear model misses.

\textbf{Online learning and adaptive policies.} Contextual bandits offer continuous policy refinement, balancing exploration and exploitation while adapting to shifting market conditions. The challenge is defining reward functions that balance target and guardrail objectives, potentially through constrained contextual bandits or safe exploration approaches.

\textbf{Long-term impact assessment.} Extended holdout designs would reveal whether treatment effects persist, dissipate, or compound over time---and whether equilibrium effects emerge as marketplace participants adapt their strategies. We are running such a holdout: a held-out portion of the marketplace continues to receive the baseline policy throughout a multi-month window while the remainder operates under the deployed personalized policy. The holdout is instrumented to (i)~measure seeker-side retention effects and (ii)~assess longer-term downstream outcomes, both of which would inform subsequent model iteration.

\section{Conclusion}
\label{sec:conclusion}

Two-sided marketplaces face a fundamental constraint: optimizing one side's experience can harm the other. Standard uplift methods assume single objectives and rich treatment variation---assumptions violated when interference necessitates cluster experiments and cross-side externalities create competing objectives. 

We developed and deployed an integrated uplift framework that addresses both challenges jointly. Our \textbf{hybrid ranking approach} separately models target and guardrail metrics, achieving over 10\% lower guardrail risk for equivalent target gains by identifying at-risk segments that single-objective optimization overlooks. Our \textbf{treatment effect extrapolation} extends CATE estimates from limited experimental levels to continuous policy optimization, with post-deployment validation confirming that linear assumptions hold within bounded ranges. The framework now serves millions of job postings, delivering statistically significant and economically sizable target lift while satisfying all engagement guardrails.

Three insights generalize beyond our setting. First, limited treatment variation need not preclude personalization: when policy decisions depend on ordinal rankings rather than precise magnitudes, linear extrapolation proves robust. Second, multi-objective tradeoffs are manageable through constrained optimization rather than contested weights---the guardrail threshold implicitly selects a point on the Pareto frontier without requiring cross-team weight negotiation. Third, ensemble CATE models address a common marketplace challenge: treatment assignment at one level (job segment) while outcomes aggregate at another (employer).

This work demonstrates that meaningful personalization is achievable in marketplace settings characterized by experimental constraints, competing objectives, and operational complexity. The methodology is applicable wherever cluster randomization limits treatment variation and multi-stakeholder tradeoffs require explicit balancing.

\begin{acks}
We thank Di Mo, Kuo-Ning Huang, and Wenjing Zhang for their support throughout this project. We are grateful to Pooja Rathnashyam, Allyson Merrigan, Jiali Huang, Dana Tom, and Alpha Bah for their collaboration on deployment. We also thank Shan Ba, Chunzhe Zhang, Sophie Sheng, and Kaixu Yang for helpful discussions.
\end{acks}

\bibliographystyle{ACM-Reference-Format}
\balance  
\bibliography{references}

\newpage
\appendix
\section{CATE Estimator Comparison}
\label{appendix:cate-comparison}

To audit sensitivity of the CATE ranking of \S\ref{sec:cate-estimation}
to meta-learner choice, we additionally implement
R-learner and generalized random forests (GRF) alongside the X-learner,
using the same training and held-out evaluation data. With only two
treatment levels, all three target the same CATE estimand
$\tau(X) = \mathbb{E}[Y(c_1) - Y(c_0) \mid X]$, so cross-method
differences in estimated rankings reflect finite-sample noise rather
than different estimands.

Cross-method Spearman rank correlations on the held-out evaluation set are
reported in Table~\ref{tab:cate-correlations}. X-learner and R-learner agree
well on both outcomes (target $\rho = 0.74$; guardrail $\rho = 0.89$),
cross-validating the X-learner choice. GRF agrees comparably on the denser
guardrail outcome ($\rho = 0.76$--$0.78$ vs.\ X and R) but diverges on the
target ($\rho = 0.35$--$0.36$). While GRF uses the same orthogonalization as
R-learner, it solves for $\tau(X)$ locally via heterogeneity-criterion tree
splits. This is consistent with local-kernel forest methods being sensitive
to per-leaf positive sparsity: under low positive-class density, local
neighborhoods lack the variation needed for stable estimates, whereas X- and
R-learners' global base-learners regularize across the sparse outcome space.

\begin{table}[!htb]
\caption{Spearman rank correlations of estimated CATE across learners on the
held-out evaluation set.}
\label{tab:cate-correlations}
\centering
\footnotesize
\begin{tabular}{@{}lccc@{}}
\toprule
\multicolumn{4}{l}{\textit{(a) Target CATE} ($\hat{\tau}_Y$)} \\
\midrule
            & X-learner & R-learner & GRF \\
\midrule
X-learner   & 1.00      & 0.74      & 0.35 \\
R-learner   & 0.74      & 1.00      & 0.36 \\
GRF         & 0.35      & 0.36      & 1.00 \\
\midrule
\multicolumn{4}{l}{\textit{(b) Guardrail CATE} ($\hat{\tau}_V$)} \\
\midrule
            & X-learner & R-learner & GRF \\
\midrule
X-learner   & 1.00      & 0.89      & 0.78 \\
R-learner   & 0.89      & 1.00      & 0.76 \\
GRF         & 0.78      & 0.76      & 1.00 \\
\bottomrule
\end{tabular}
\end{table}

\section{CATE Estimation Noise and Ranking Stability}
\label{appendix:bootstrap-stability}

\begin{figure}[H]
    \centering
    \captionsetup[subfigure]{font=footnotesize}
    \begin{subfigure}[b]{0.48\columnwidth}
        \includegraphics[width=\textwidth]{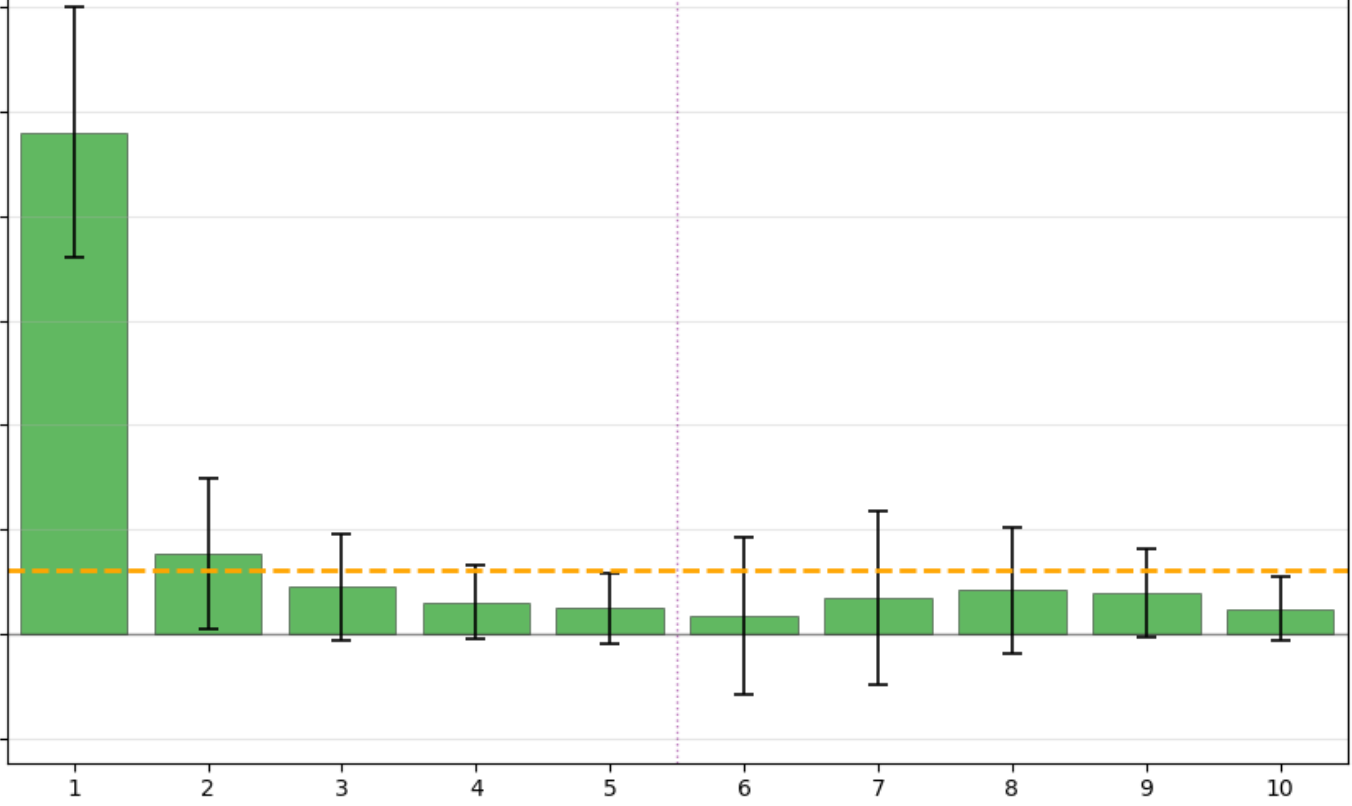}
        \caption{Holdout Target Effect by Hybrid Ranking}
    \end{subfigure}
    \hfill
    \begin{subfigure}[b]{0.48\columnwidth}
        \includegraphics[width=\textwidth]{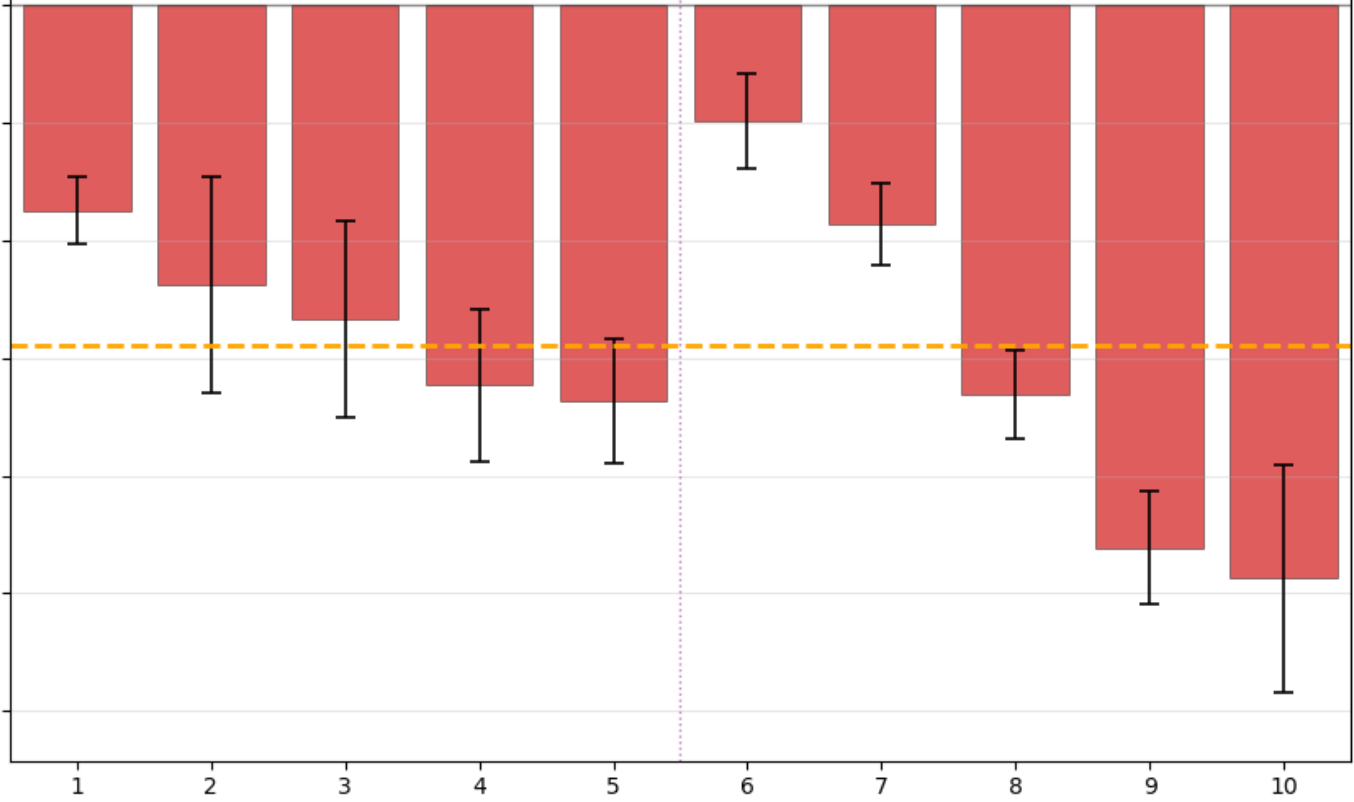}
        \caption{Holdout Guardrail Effect by Hybrid Ranking}
    \end{subfigure}
    \caption{Holdout Treatment Effect per Bootstrapped Hybrid Ranking}
    \label{fig:bootstrap-hybrid-ate}
    \Description{Two-panel bar chart with confidence-interval whiskers showing
    cross-bootstrap-replicate mean observed treatment effects on target (left)
    and guardrail (right) per rep-defined hybrid ranking group, with broadly
    monotonic ordering consistent with the hybrid ranking design.}
    \vspace{-1ex}
    {\small\textit{Note:} Orange dashed line indicates ATE benchmark.}
\end{figure}

To verify that the hybrid ranking group assignment is robust to
CATE estimation noise, we re-estimate the full X-learner on $B=100$
cluster-stratified bootstrap resamples of the training set.
For each replicate $b$, we recompute the per-job CATE predictions
$\hat{\tau}_Y^{(b)}, \hat{\tau}_V^{(b)}$ on the held-out evaluation set, assign
each job to its rep-specific hybrid ranking group following
\S\ref{sec:hybrid-ranking}, and aggregate observed treatment effects per group.
Figure~\ref{fig:bootstrap-hybrid-ate} shows the cross-replicate mean (bars) and
90\% confidence interval (whiskers) for both panels.
Within the top-50\% groups (1--5) the target panel exhibits broadly monotonic
decline consistent with the target-CATE sub-ranking; within the bottom-50\%
groups (6--10) the guardrail panel similarly exhibits broadly monotonic decline
consistent with the guardrail-CATE sub-ranking. The confidence interval bounds remain
broadly monotonic across the hybrid grouping (for target in the top half and for guardrail in the bottom half), indicating the production
hybrid-group assignment is stable under CATE estimation noise.

\section{Algorithm Pseudocode}

Due to proprietary constraints, we cannot release experimental data. The methodology is described in sufficient detail for application to comparable marketplace data.

\begin{algorithm}[H]
\caption{Multi-Objective Hybrid Ranking Policy Assignment}
\label{alg:hybrid}
\begin{algorithmic}[1]
\REQUIRE Job features $X$, experimental data $D$, guardrail threshold $\delta$
\ENSURE Policy assignment $\pi(X)$ for each job
\STATE Estimate CATE models $\hat{\tau}_Y$, $\hat{\tau}_V$ using X-learner on $D$
\FOR{each job $j$}
    \STATE Compute $\hat{\tau}_Y(X_j)$: target metric uplift
    \STATE Compute $\hat{\tau}_V(X_j)$: engagement impact
\ENDFOR
\STATE Compute extrapolated effects for free-value thresholds $c$
\STATE Create hybrid ranking:
\STATE \quad Sort jobs by $\hat{\tau}_Y$ descending for top 50\%
\STATE \quad Sort remaining jobs by $\hat{\tau}_V$ descending (least negative first)
\STATE Assign tiered policies based on hybrid ranking position
\STATE Verify: $\sum_j \hat{\tau}_V(\pi(X_j)) \geq -\delta$
\IF{constraint violated}
    \STATE Relax policies for lowest-ranked jobs; repeat verification
\ENDIF
\RETURN Policy assignments $\pi$
\end{algorithmic}
\end{algorithm}

\end{document}